# A Stochastic Gradient Method with Biased Estimation for Faster Nonconvex Optimization

Jia Bi, Steve R.Gunn

School of Electronics and Computer Science, University of Southampton, United Kingdom
jb4e14@soton.ac.uk, srg@ecs.soton.sc.uk

## Abstract

A number of optimization approaches have been proposed for optimizing nonconvex objectives (e.g. deep learning models), such as batch gradient descent, stochastic gradient descent and stochastic variance reduced gradient descent. Theory shows these optimization methods can converge by using an unbiased gradient estimator. However, in practice biased gradient estimation can allow more efficient convergence to the vicinity since an unbiased approach is computationally more expensive. To produce fast convergence there are two trade-offs of these optimization strategies which are between stochastic/batch, and between biased/unbiased. This paper proposes an integrated approach which can control the nature of the stochastic element in the optimizer and can balance the trade-off of estimator between the biased and unbiased by using a hyper-parameter. It is shown theoretically and experimentally that this hyper-parameter can be configured to provide an effective balance to improve the convergence rate.

## 1 Introduction

Optimizing methods for nonconvex problems has become a crucial research topic in artificial intelligence, e.g. deep neural networks, graphs. The objective function for the parameter optimization can be formulated as a finite-sum problem,

$$\min_{x \in \mathbb{R}^d} f(x), \qquad f(x) := \frac{1}{n}\sum_{i=1}^{n} f_i(x), \qquad (1)$$

where the individual $f_i (i \in [n])$ and $f$ are nonconvex but Lipschitz smooth ($\mathcal{L}$-smooth) [Strongin and Sergeyev, 2000; Reddi *et al.*, 2016b]. We use $\mathcal{F}_n$ to denote all functions of Eq. 1 and analyse our optimization method using the Incremental First-order Oracle (IFO) framework [Reddi *et al.*, 2016b; Agarwal and Bottou, 2015]. Based on complexity analysis, IFO is a way to evaluate lower bounds for finite-sum problems [Bottou *et al.*, 2018; Johnson and Zhang, 2013]. The underlying training algorithms for nonconvex problems are still stochastic gradient descent (SGD) and its heuristic variants to solve Eq. 1 [Allen-Zhu, 2018]. One of the variants is variance reduced (VR) based stochastic optimization approaches (e.g. stochastic variance reduced gradient (SVRG)) which has been shown to accelerate the convergence rate of SGD by reducing the noise of gradients on nonconvex problems [Johnson and Zhang, 2013]. The SVRG algorithm is showed in Alg. 1 [Johnson and Zhang, 2013]. However, VR-based stochastic algorithms have three

---

**Algorithm 1:** SVRG $(x^0, \eta, \{p_i\}_{i=0}^m, m, S)$

**Input:** Epoch length m, learning rate $\eta$, number of epochs $S = T/m$ where T is total number of iterations, discrete probability distribution $\{p_i\}_{i=0}^m$;

1 Initialize $\tilde{x}^0 = x_m^0 = x^0$;
2 **for** s = 0 **to** S − 1 **do**
3 $\quad x_0^{s+1} = x_m^s$; $g^{s+1} = \frac{1}{n}\sum_{i=1}^n \nabla f_i(\tilde{x}^s)$;
4 $\quad$ **for** t = 0 **to** m − 1 **do**
5 $\quad\quad$ Randomly select $i_t$ from $\{1, ..., n\}$ ;
6 $\quad\quad v_t^{s+1} = \nabla f_{i_t}(x_t^{s+1}) - \nabla f_{i_t}(\tilde{x}^s) + g^{s+1}$;
7 $\quad\quad x_{t+1}^{s+1} = x_t^{s+1} - \eta v_t^{s+1}$;
8 $\quad \tilde{x}^{s+1} = \sum_{i=0}^m p_i x_i^{s+1}$;

**Output:** $\tilde{x}^S$

---

problems. Firstly, VR schemes reduce the ability to escape local minima in later iterations due to a diminishing variance. The challenge in VR-based stochastic optimization is therefore to control the reduction in variance. Secondly, SVRG is a unbiased estimation, which is not efficient to be used in practice so as to be replaced by biased estimation. There are various reasons, e.g. because unbiased estimator is difficult to compute [Liang *et al.*, 2009] or because biased estimation can give a lower upper bound of mean squared error (MSE) loss function. Thirdly, the learning rate in such an algorithm is fixed and relatively large, which has the advantage of encouraging initial points out of local minima in early iterations but can hinder optimization convergence to a local optima.

Table 1: Compare the IFO complexity of different algorithms. Generally, the schedule of learning rate in SGD is decayed by increasing number of iteration. In SVRG (Alg. 1), the learning rate is fixed. MSVRG [Reddi *et al.*, 2016b] is a modified SVRG method as the method modify the learning rate that in each iteration is chose the maximum value of two items, including the decayed value by the increasing number of iterations, and the other fixed value by the number of training samples. For ISVRG, same adaptive learning rate with MSVRG.

| Algorithm | IFO calls on Nonconvex | The schedule of Learning rate $\eta$ |
|---|---|---|
| SGD | $\mathcal{O}(1/\varepsilon^2)$ | Decayed |
| SVRG | $\mathcal{O}(n + (n^{2/3}/\varepsilon))$ | Fixed |
| MSVRG | $\mathcal{O}(\min(1/\varepsilon^2, n^{2/3}/\varepsilon))$ | max{Decayed, Fixed} |
| ISVRG$^+$ | $\mathcal{O}(\min(1/\varepsilon^2, \mathbf{n^{1/5}}/\varepsilon))$ | max{**Decayed, Fixed**} |

To address these three problems, we propose our method *Integrated biased SVRG* (ISVRG$^+$) which can control the reduced variance and choose the biased or unbiased estimator in each iteration to accelerate the convergence rate of nonconvex optimization.

**Contributions** We summarize and list our main contributions:

- We introduce ISVRG$^+$, a well-balanced VR method for SGD. We provide a theoretical analysis of our algorithm on nonconvex problems.
- ISVRG$^+$ balances the trade-off between biased and unbiased estimation, which can provide a fast convergence rate.
- Compared with SGD and SVRG-based optimization, our method can achieve comparable or faster rates of convergence. To the best of our knowledge, we provide the first analysis about controlling the variance reduction to balance the gradient of SVRG and balance the nature of estimator between biased and unbiased to obtain provably superior performance to SGD and its variants on nonconvex problems. Table 1 compares the theoretical rates of convergence of four methods, which shows that ISVRG$^+$ has the fastest rate of convergence.
- We show empirically that ISVRG$^+$ has faster rates of convergence than SGD, SVRG and MSVRG [Reddi *et al.*, 2016b] which is a modified SVRG using an adaptive learning schedule with standard SVRG.

## 2 Preliminaries

For our analysis, we require the following background to introduce definitions for $\mathcal{L}$-smooth, $\varepsilon$-accuracy, and $\sigma$-bounded gradients. We assume the individual functions $f_i$ in Eq. 1 are $\mathcal{L}$-smooth which is to say that

$$\| \nabla f_i(x) - \nabla f_i(y) \| \leq L \| x - y \|, \quad \forall x, y \in \mathbb{R}^d. \quad (2)$$

An $\varepsilon$-accuracy criterion can be used to analyse the convergence of nonconvex forms of Eq. 1 [Ghadimi and Lan, 2016].

**Definition 1.** *A point* $x$ *is called* $\varepsilon$-*accurate if* $\| \nabla f(x)^2 \| \leq \varepsilon$. *An iterative stochastic algorithm can achieve* $\varepsilon$-*accuracy within* $t$ *iterations if* $\mathbb{E}[\| \nabla f(x^t) \|^2] \leq \varepsilon$, *where the expectation is over the stochastic algorithm.*

**Definition 2.** $f \in \mathcal{F}_n$ *has* $\sigma$-*bounded gradient if* $\| \nabla f_i(x) \| \leq \sigma$ *for all* $i \in [n]$ *and* $x \in \mathbb{R}$.

The following theorems showed two upper bounds of expectation of standard SVRG responding to two cases of learning rate. Particularly, to fairly compare the upper bound between our methods (introduced in later sections) and standard SVRG, we rescale the standard SVRG by multiplying 0.5 to the gradient $v_t^{s+1}$ in Alg 1. In the first case of the learning rate depending on iteration number T, we use $\sigma$-bounded gradients showed in Definition 2 to achieve the upper bound as following.

**Theorem 1.** *Suppose* $f$ *has* $\sigma$-*bounded gradients. Let* $\eta_{t_{SVRG}} = \eta_{SVRG} = C_{SVRG}/\sqrt{T}$ *where* $C_{SVRG} = \sqrt{\dfrac{f(x^0) - f(x^*)}{2L\sigma^2}}$, *and* $x^*$ *is an optimal solution to Eq. 1. Alg. 1 satisfies*

$$\min_{0 \leq t \leq T-1} \mathbb{E}[\| \nabla f(x^t) \|^2] \leq \sqrt{2}\sqrt{\dfrac{2(f(x^0) - f(x^*))L}{T}}\sigma.$$

In the second case of learning rate depending on the training sample size n, we can achieve a upper bound without $\sigma$-bounded is shown in following theorem.

**Theorem 2.** *Let* $f \in \mathcal{F}_n$, *let* $c_{m_{SVRG}} = 0$, $\eta > 0$, $\beta_t = \beta > 0$ *and* $c_{t_{SVRG}} = c_{t+1}(1+\eta\beta+2\eta^2L^2)+L^3\eta^2$, *so the intermediate result* $\Omega_{t_{SVRG}} = (\eta - \dfrac{c_{t+1}\eta}{\beta_t} - L\eta^2 - 2c_{t+1}\eta^2) > 0$, *for* $t$ *from 0 to* $m - 1$. *Define the minimum value of* $\gamma_{n_{SVRG}} := \min_t \Omega_{t_{SVRG}}$. *Further let* $p_i = 0$ *where* $0 \leq i < m$, $p_m = 1$, *and* T *is a multiple of* m. *Defining the output of Alg. 1 as* $x_a$ *we have the following upper bound:*

$$\mathbb{E}[\| \nabla f(x_a) \|^2] \leq \dfrac{f(x^0) - f(x^*)}{T\gamma_{n_{SVRG}}},$$

*where* $x^*$ *is an optimal solution to Eq. 1.*

Further, to achieve an explicit upper bound in Theorem 2 it is necessary to define the relationship between $\gamma_{n_{SVRG}}$ and n. we specify $\eta$ and $\beta$ following [Reddi *et al.*, 2016b; Reddi *et al.*, 2016c], resulting in the following theorem.

**Theorem 3.** *Suppose* $f \in \mathcal{F}_n$, $\eta = \dfrac{1}{3Ln^\alpha}$ ($0 < \mu_0 < 1$ *and* $0 < \alpha \leq 1$), $\beta = L/n^{\alpha/2}$, $m_{SVRG} = \lfloor \dfrac{9n^{\alpha/2}}{5} \rfloor$, T *is the total number of iterations which is a multiple of* $m_{SVRG}$, *and*

$v_{SVRG} > 0$. So we have $\gamma_{n_{SVRG}} \geq \frac{v_{SVRG}}{18Ln^\alpha}$ in Theorem 2. The output $x_a$ of Alg. 1 satisfies

$$\mathbb{E}[\| \nabla f(x_a) \|^2] \leq \frac{18Ln^\alpha[f(x^0) - f(x^*)]}{Tv_{SVRG}},$$

where $x_*$ is an optimal solution to Eq. 1.

## 3 Related works

Many SGD-based methods have been applied to optimize functions in different domains. For example, one approach is stochastic average gradient (SAG), which uses a memory of previous gradient values to achieve a linear convergence rate, which can be used to optimize a finite set of smooth functions in a strongly-convex domain [Schmidt *et al.*, 2013]. Further, inspired from SAG and SVRG, SAGA is an incremental gradient algorithm with fast linear convergence rate that can be used in three different domains, including non-strongly convex problems [Defazio *et al.*, 2014], nonconvex but linear problems [Reddi *et al.*, 2016a]. For non-smooth nonconvex finite-sum functions, proximal operators to handle nonsmoothness in convex problem can cooperate with nonconvex optimization as ProxSGD, ProxGD, ProxSAG, ProSVRG, ProxSAGA and so on. [Reddi *et al.*, 2016c; Sra, 2012].

In some real-world learning scenario requiring expensive computation of the sample gradient, e.g. graph deep learning models, an unbiased estimator is usually computationally expensive or not available [Chen and Luss, 2018]. As a result, many works have proposed asymptotically biased optimizations with biased gradient estimators as economic alternative to an unbiased version that does not converge to the minima, but to their vicinity [Chen *et al.*, 1987; fu Chen and Gao, 1989; Chen and Luss, 2018; Tadić and Doucet, 2017; Chen *et al.*, 2018]. These methods provide a good insight into the biased gradient search, however they hold under restrictive conditions which are very hard to verify for complex stochastic gradient algorithms. In this paper, we analyse the nature of biased and unbiased estimators in different stages of the optimization process on nonconvex problems, and proposed a method combining the benefits of both biased and unbiased estimator to achieve a fast convergence rate.

## 4 Integrated SVRG with biased estimation

For the first challenge of SVRG, the balance of the gradient update between the full batch and stochastic estimators is fixed. We introduces a hyper-parameter $\lambda$ to balance the weighting of the stochastic element with the full batch gradient to allow the algorithm to choose appropriate behaviours from stochastic, through reduced variance, to batch gradient descent. As a result, the adoption of $\lambda$ in the first-order iterative algorithm can gain benefits from the stochastic estimator to speed-up computation and escape the local minimum, and reduced variance to accelerate the rates of convergence.

To address the second challenge associated with the trade-off between biased/unbiased estimator, we still consider to use this hyper-parameter $\lambda$ to choose the appropriate estimator from biased to unbiased during the whole optimization.

In terms of the third challenge of SVRG associated with fixed learning rates, some research has shown that adaptive learning rates can be applied with reduced variance to provide faster convergence rates on nonconvex optimization [Reddi *et al.*, 2016b; Goodfellow *et al.*, 2016]. We followed the work of MSVRG method [Reddi *et al.*, 2016b], the adaptive learning schedule is chosen to maximize between two cases of learning schedules which are based on the increasing number of iterations $t$ and the number of samples $n$. Thus, the learning rate can be decayed by increasing the number of iterations but is also lower bounded by the data size to prevent the adaptive learning rate from decreasing too quickly.

---

**Algorithm 2:** $SVRG_{unbiased}(x^0, \{\eta_i\}_{i=0}^T, \{p_i\}_{i=0}^m, m, S)$

**Input**: Same input parameters with Alg 1;
1 Initialize $\tilde{x}^0 = x_m^0 = x^0$;
2 **for** *s=0 to S-1* **do**
3 $\quad x_0^{s+1} = x_m^s$; $g^{s+1} = \frac{1}{n}\sum_{i=1}^n \nabla f_i(\tilde{x}^s)$;
4 $\quad$ **for** t = 0 **to** m - 1 **do**
5 $\quad\quad$ Randomly select $i_t$ from $\{1,...,n\}$ ;
6 $\quad\quad v_t^{s+1} = (1-\lambda)\nabla f_{i_t}(x_t^{s+1}) - \lambda\left(\nabla f_{i_t}(\tilde{x}^s) - g^{s+1}\right)$;
7 $\quad\quad x_{t+1}^{s+1} = x_t^{s+1} - \eta_\Delta v_t^{s+1}$;
8 $\quad \tilde{x}^{s+1} = \sum_{i=0}^m p_i x_i^{s+1}$;
**Output**: $\tilde{x}^S$

---

### 4.1 Weighted unbiased estimator analysis

Standard SVRG is unbiased showed in Alg1. In this section, we introduce a weighted standard SVRG shown in Alg 2, which can weight the terms of stochastic and batch by $\lambda$. Under appropriate conditions, we can achieve results including the following theorem.

**Theorem 4.** *Suppose* $f \in \mathcal{F}_n$ *have* $\sigma$-*bounded gradient. Let* $\eta_t = \eta_{\Delta_{unbiased}} = c_{unbiased}/\sqrt{\Delta+1}$ *for* $0 \leq \Delta \leq T-1$ *where* $c_{unbiased} = \sqrt{\frac{f(x_0) - f(x^*)}{(2\lambda^2 - 2\lambda + 1)L\sigma^2}}$ *and let* $T$ *be a multiple of* $m$. *Further let* $p_m = 1$, *and* $p_i = 0$ *for* $0 \leq i < m$. *Then the output* $x_a$ *of Alg. 2 we have*

$$\mathbb{E}[\| \nabla f(x_a)^2 \|] \leq \frac{\sqrt{(2\lambda^2 - 2\lambda + 1)}}{(1-\lambda)}\sqrt{\frac{2(f(x^0) - f(x^*))L}{T}}\sigma$$

In Theorem 4, we schedule a decayed learning rate in this theorem $\eta_{\Delta_{unbiased}} \propto 1/\sqrt{\Delta+1}$ [Reddi *et al.*, 2016b], which can avoid to know the total number of inner iterations across all epochs $T$ in Alg. 2 in advance. Compared with the upper bound in Theorem 1, we can achieve a lower upper bound in Theorem 4 if $0 \leq \lambda < \frac{1}{2}$ and the optimal value of $\lambda^* = 0$.

In the second case the learning rate $\eta_t$ is fixed depending upon data size $n$, we can achieve the results in the following theorems.

**Theorem 5.** *Let* $f \in \mathcal{F}_n$, *let* $c_m = 0$, $\eta_t = \eta > 0$, $\beta_t = \beta > 0$, $c_{t_{unbiased}} = c_{t+1}(1 + (1-\lambda)\eta\beta + 2(1-\lambda)^2\eta^2L^2) + L^3\eta^2$, *so the intermediate result* $\Omega_{t_{unbiased}} = (\eta_t - (1-\lambda)\frac{c_{t+1}\eta_t}{\beta_t} - (1-\lambda)^2 L\eta_t^2 - 2(1-\lambda)^4 c_{t+1}\eta_t^2) > 0$, *for* $0 \leq t \leq m-1$. *Define the minimum value of* $\gamma_{n_{unbiased}} := \min_t \Omega_{t_{unbiased}}$. *Further let* $p_i = 0$ *for* $0 \leq i < m$ *and* $p_m = 1$, *and* $T$ *is a multiple of* $m$. *So the output* $x_a$ *of Alg. 2 we have*

$$\mathbb{E}[\| \nabla f(x_a) \|^2] \leq \frac{f(x^0) - f(x^*)}{T\gamma_{n_{unbiased}}},$$

*where* $x^*$ *is the optimal solution to Problem 1.*

We use the same schedule of $\eta$ and $\beta$ as in Theorem 3 to determine $\gamma_{n_{unbiased}}$ in the following theorems.

**Theorem 6.** *Suppose* $f \in \mathcal{F}_n$, *let* $\eta = \frac{1}{3Ln^{a\alpha}}$ *($0 \leq a \leq 1$, and $0 < \alpha \leq 1$)*, $\beta = \frac{L}{n^{b\alpha}}$ *($b > 0$)*, $m_{unbiased} = \lfloor \frac{3n^{(3a+b)\alpha}}{(1-\lambda)} \rfloor$ *and* $T$ *is the total number of iterations. Then, we can obtain the lower bound* $\gamma_{n_{unbiased}} \geq \frac{(1-\lambda)\nu}{9n^{(2a-b)\alpha}L}$ *in Theorem 5. For the output* $x_a$ *of Alg. 2 we have*

$$\mathbb{E}[\| \nabla f(x_a) \|^2] \leq \frac{9n^{(2a-b)\alpha}L[f(x^0) - f(x^*)]}{(1-\lambda)T\nu},$$

*where* $x_*$ *is an optimal solution to Eq. 1.*

where the $\lambda \neq 0$ since the gradient in this case is no $\sigma$-bound. As a result, compare with Theorem 3, we can achieve a lower upper bound in above theorem when $0 < \lambda < 1 - \frac{n^{(2a-b-1)\alpha}}{2} < 1$ and the optimal value of $\lambda^* \to 0$.

### 4.2 Biased estimator analysis

In this section we theoretically analyse the performance of biased SVRG using same learning rate schedule with weighted unbiased version, which the algorithm shown in Alg 3. In the first case of learning rate, biased SVRG can use $\sigma$-bounded gradient in Definition 2 when satisfying a condition that $0 \leq \lambda \leq \frac{2}{3}$. We provide a proof for this condition.

*Proof.* As the learning rate decay from 1 to T, we use Definition 2 to bound gradients $v_t^{s+1}$ as following:

$\mathbb{E}[\| v_t^{s+1} \|^2]$
$= \mathbb{E}[\| (1-\lambda)(\nabla f_{i_t}(x_t^{s+1}) - \nabla f_{i_t}(\tilde{x}^s)) + \lambda\nabla f(\tilde{x}^s) \|^2]$
$= \mathbb{E}[\| (1-\lambda)\nabla f_{i_t}(x_t^{s+1}) - (1-\lambda)\nabla f_{i_t}(\tilde{x}^s) + \lambda\nabla f(\tilde{x}^s) \|^2]$
$\leq 2(\mathbb{E}[(\| (1-\lambda)\nabla f_{i_t}(x_t^{s+1}) \|^2 + \| (1-\lambda)\nabla f_{i_t}(\tilde{x}^s) - \lambda\nabla f(\tilde{x}^s) \|^2])$
$\leq 2((1-\lambda)^2\mathbb{E}[\| \nabla f_{i_t}(x_t^{s+1}) \|^2] + (1-\lambda)^2\mathbb{E}[\| \nabla f_{i_t}(\tilde{x}^s) \|^2])$
$\leq 4(1-\lambda)^2\sigma^2,$
(3)

---

**Algorithm 3:** $SVRG_{biased}(x^0, \{\eta_i\}_{i=0}^T, \{p_i\}_{i=0}^m, m, S)$

**Input**: Same input parameters with Alg 1
1  Initialize $\tilde{x}^0 = x_m^0 = x^0$;
2  **for** *s=0 to S-1* **do**
3  $\quad x_0^{s+1} = x_m^s$; $g^{s+1} = \frac{1}{n}\sum_{i=1}^n \nabla f_i(\tilde{x}^s)$;
4  $\quad$ **for** t = 0 to m − 1 **do**
5  $\quad\quad$ Randomly select $i_t$ from $\{1, ..., n\}$;
6  $\quad\quad$ $v_t^{s+1} = (1-\lambda)\left(\nabla f_{i_t}(x_t^{s+1}) - \nabla f_{i_t}(\tilde{x}^s)\right) + \lambda g^{s+1}$;
7  $\quad\quad$ $x_{t+1}^{s+1} = x_t^{s+1} - \eta_\Delta v_t^{s+1}$;
8  $\quad \tilde{x}^{s+1} = \sum_{i=0}^m p_i x_i^{s+1}$;
**Output**: $\tilde{x}^S$

where tche second inequality we followed (a) $\sigma$-bounded gradient property of $f$ and (b) for a biased random variable $\zeta$ which has a upper bounding as

$\mathbb{E}[\| (1-\lambda)\zeta - \lambda\mathbb{E}[\zeta] \|^2]$
$= \mathbb{E}[(1-\lambda)^2 \| \zeta \|^2 - 2(1-\lambda)\lambda\zeta\mathbb{E}[\zeta] + \lambda^2\mathbb{E}^2[\zeta]]$
$= (1-\lambda)^2\mathbb{E}[\| \zeta \|^2] - (2\lambda - 3\lambda^2)\mathbb{E}^2[\zeta] \leq (1-\lambda)^2\mathbb{E}[\| \zeta \|^2],$
(4)

where the upper bound requires to satisfy the fact that for a unbiasedness of random variable $\zeta$, $\mathbb{E}[\| (1-\lambda)^2\zeta - (1-\lambda)^2\mathbb{E}[\zeta] \|^2] < \mathbb{E}[\| (1-\lambda)^2\zeta \|^2]$. Thus, the $\lambda$ should be within the range that $0 \leq \lambda \leq \frac{2}{3}$. □

We then achieved results from the following theorem.

**Theorem 7.** *Suppose* $f \in \mathcal{F}_n$ *have $\sigma$-bounded gradient. Let* $\eta_{t_{biased}} = \eta_\Delta = c_{biased}/\sqrt{\Delta + 1}$ *for* $0 \leq \Delta \leq T-1$ *where* $c_{biased} = \sqrt{\frac{f(x_0) - f(x^*)}{2\lambda L\sigma^2}}$ *and let* $T$ *be a multiple of* $m$. *Further let* $p_m = 1$, *and* $p_i = 0$ *for* $0 \leq i < m$. *Then the output* $x_a$ *of Alg. 3 we have*

$$\mathbb{E}[\| \nabla f(x_a)^2 \|] \leq \frac{2(1-\lambda)}{\sqrt{\lambda}}\sqrt{\frac{2(f(x^0) - f(x^*))L}{T}}\sigma$$

Consequently, we can achieve a lower upper bound of expectation in Theorem 7 than scaled standard SVRG in Theorem 1 when $\lambda$ satisfying the two ranges simultaneously that $0 \leq \lambda \leq \frac{2}{3}$ and $\frac{1}{2} < \lambda \leq 1$. Consequently, the range of $\lambda$ is $\frac{1}{2} < \lambda \leq \frac{2}{3}$, and the optimal value of $\lambda = \lambda^* = \frac{2}{3}$.

In terms of the second case of learning rate, the biased version SVRG can obtain its upper bound of expectation shown in the following theorems.

**Theorem 8.** *Let* $f \in \mathcal{F}_n$, *let* $c_m = 0$, $\eta_t = \eta > 0$, $\beta_t = \beta > 0$, $c_{t_{biased}} = c_{t+1}(1 + \eta\beta + 2(1-\lambda)^2\eta^2L^2) + L^3\eta^2(1-\lambda)^2$, *so the intermediate result* $\Omega_{t_{biased}} = (\eta_t - \frac{c_{t+1}\eta_t}{\beta_t} - \lambda^2 L\eta_t^2 - 2\lambda^2 c_{t+1}\eta_t^2) > 0$, *for* $0 \leq t \leq m-1$. *Define the minimum value of* $\gamma_n := \min_t \Omega_{t_{biased}}$. *Further let* $p_i = 0$ *for* $0 \leq i < m$ *and*

$p_m = 1$, and T is a multiple of m. So the output $x_a$ of Alg. 3 we have

$$\mathbb{E}[\| \nabla f(x_a) \|^2] \leq \frac{f(x^0) - f(x^*)}{T\gamma_{n_{biased}}},$$

where $x^*$ is the optimal solution to Problem 1.

To determine $\gamma_{n_{biased}}$, we use same schedule of $\eta$ and $\beta$ from Theorem 3 and 5 and then achieve result in the following theorem.

**Theorem 9.** Suppose $f \in \mathcal{F}_n$, let $\eta = \frac{1}{3Ln^{a\alpha}}$ ($0 \leq a \leq 1$ and $0 < \alpha \leq 1$), $\beta = \frac{L}{n^{b\alpha}}$ ($b > 0$), $m_{biased} = \lfloor \frac{3n^{2a\alpha}}{2(1-\lambda)} \rfloor$ and T is the total number of iterations. Then, we can obtain the lower bound $\gamma_{n_{biased}} \geq \frac{(1-\lambda)\lambda v_1}{9Ln^{(2a-b)\alpha}}$ in Theorem 8. For the output $x_a$ of Alg. 3 we have

$$\mathbb{E}[\| \nabla f(x_a) \|^2] \leq \frac{9Ln^{(2a-b)\alpha}[f(x^0) - f(x^*)]}{\lambda(1-\lambda)Tv_1},$$

where $x_*$ is an optimal solution to Eq. 1.

Where the $\lambda \neq 0$ since the gradient is no $\sigma$-bound. Compared with the upper bound of expectation in Theorem 3, we can achieve a lower upper bound when $\lambda$ satisfying $0 < \frac{1 - \sqrt{1 - 4n^{(2a-b-1)\alpha}}}{2} < \lambda < \frac{1 + \sqrt{1 - 4n^{(2a-b-1)\alpha}}}{2} < 1$. The optimal value of $\lambda^* = 0.5$.

### 4.3 Variance control and combined biased and unbiased estimation

To estimate the performance of unbiased and biased estimators, we compared the value of upper bound depending on two cases of learning rate. Particularly, in the first case the learning rate decayed by iteration number, the upper bound of biased version when $\lambda^* = 2/3$ in Theorem 4 is lower than unbiased version when $\lambda^* = 0$ in Theorem 7. This result shows that the biased and weighted SVRG estimator can provide a lower upper bound than unbiased SGD when learning rate is decayed. And in the second case the learning rate fixed by training samples, the upper bound of unbiased version when $\lambda^* \to 0$ in Theorem 5 is lower than biased version when $\lambda^* = 0.5$ in Theorem 8. Besides, in this case of learning rate is fixed, standard SVRG is better than SGD. So it shows that unbiased and weighted SVRG is better than biased standard SVRG and SGD.

Consequently, these results give rise to a new optimization method proposed by $\lambda$ which not only can combine biased and unbiased SVRG but also can control the reduced variance so as to improve the rates of convergence, which the new method as **ISVRG⁺** is shown in Alg 4. Followed Alg 4, we can achieve general result for **ISVRG⁺** in the following theorem.

**Theorem 10.** Let $f \in \mathcal{F}_n$ have $\sigma$-bounded gradients. Let $\eta_t = \eta_\Delta = \max\left\{\frac{c}{\sqrt{\Delta + 1}}, \frac{1}{3Ln^{a\alpha}}\right\}$ for $\Delta$ from 0 to $T - 1$,

---

**Algorithm 4:** ISVRG⁺($x^0, \{\eta_i\}_{i=0}^T, \{p_i\}_{i=0}^m, m, S$)

**Input** : Same input parameters with Alg 1, and learning rate $\eta_s = \max\{\frac{\sqrt{c_{biased}}}{ms}, \frac{1}{3Ln^{a\alpha}}\}$, $\lambda \to 0$;

1 Initialize $\tilde{x}^0 = x_m^0 = x^0$;
2 **for** s=0 **to** S-1 **do**
3     $x_0^{s+1} = x_m^s$; $g^{s+1} = \frac{1}{n}\sum_{i=1}^n \nabla f_i(\tilde{x}^s)$;
4     **for** t = 0 **to** m − 1 **do**
5        Randomly select $i_t$ from $\{1, ..., n\}$ ;
6        **if** $\eta_s = \frac{\sqrt{c_{biased}}}{ms}$ **then**
7           $v_t^{s+1} = \frac{1}{3}\left(\nabla f_{i_t}(x_t^{s+1}) - \nabla f_{i_t}(\tilde{x}^s)\right) + \frac{2}{3}g^{s+1}$
8        **else if** $\eta_s = \eta = \frac{1}{3Ln^{a\alpha}}$ **then**
9           $v_t^{s+1} = (1-\lambda)\nabla f_{i_t}(x_t^{s+1}) - \lambda\left(\nabla f_{i_t}(\tilde{x}^s) - g^{s+1}\right)$
10       $x_{t+1}^{s+1} = x_t^{s+1} - \eta_\Delta v_t^{s+1}$;
11     $\tilde{x}^{s+1} = \sum_{i=0}^m p_i x_i^{s+1}$;

---

$m = \lfloor n^{(3a+b)\alpha} \rfloor$, and $c = \sqrt{\frac{3f(x^0) - f(x^*)}{4L\sigma^2}}$. Further let T is a multiple of m, $p_m = 1$ and $p_i = 0$ for $0 \leq i < m$. Then, the output $x_a$ of Alg. 4 satisfies

$$\mathbb{E}[\| \nabla f(x_a) \|^2]$$
$$\leq \tilde{v} \min\{\sqrt{3}\sqrt{\frac{(f(x^0) - f(x^*))L}{T}}\sigma, \frac{9n^{(2a-b)\alpha}L[f(x^0) - f(x^*)]}{Tv_2}\},$$

where $x_*$ is an optimal solution to Eq. 1, $0 \leq a \leq 1$, $0 < \alpha \leq 1$ and $b > 0$. $\tilde{v}$, $v_2$ are universal constants.

We specify the optimal value of parameters including a, b and $\alpha$ from Theorem 10. Such parameters can give rise to the following key result of the paper, which are showed in Corollary 1. In this corollary, the IFO complexity of **ISVRG⁺** is the minima between $1/\varepsilon^2$ which is equal to the IFO complexity of SGD method [Reddi et al., 2016b; Ghadimi and Lan, 2016] and $n^{1/5}/\varepsilon$ where the optimal value of a, b and $\alpha$ can be found by two conditions: (a) the upper bound of **ISVRG⁺** is lower than that of scaled standard SVRG in Theorem 3 on magnitude level as $n^0 < n^{(2a-b)\alpha} \leq n^\alpha$, if $\lambda = \lambda^* = 0$. (2) $0 < \alpha = 1/(3a + b) \leq 1$ when $0 \leq a \leq 1$ and $b > 0$. Thus, we can achieve lowest upper bound when $a = 1$, $b = 2$ and $\alpha = 1/5$. As a result, our result is possible to be better than SGD if the $\varepsilon$ within 0 to $1/n^{1/5}$.

**Corollary 1.** Suppose $f \in \mathcal{F}_n$, the IFO complexity of Alg. 4 (with parameters from Theorem 10) achieves an $\varepsilon$-accurate solution that is $\mathcal{O}(\min\{1/\varepsilon^2, n^{1/5}/\varepsilon\})$, where the number of IFO calls is minimized when $a = 1$, $b = 2$ and $\alpha = 1/5$.

Following the range of $\lambda$ obtained by above theorems, we now discuss how the hyper-parameter $\lambda$ works with the hybrid adaptive learning rate to control the variance and to balance the trade-off between bias/unbiased estimation. For the case where the first term of the learning rate is larger than second term when t is small, larger $\lambda$ with biased estimator can reduce variance and provide a lower upper bound than unbiased version, which can accelerate the convergence in early iterations. On the other hand, when the second term becomes larger later in optimization and being stable, the value of $\lambda \to 0$ with unbiased estimator allow gradient behave more stochastically. Higher variance will help gradients escape from local minimal. Meanwhile, the estimator is unbiased, which can guarantee to find the objective point in the end.

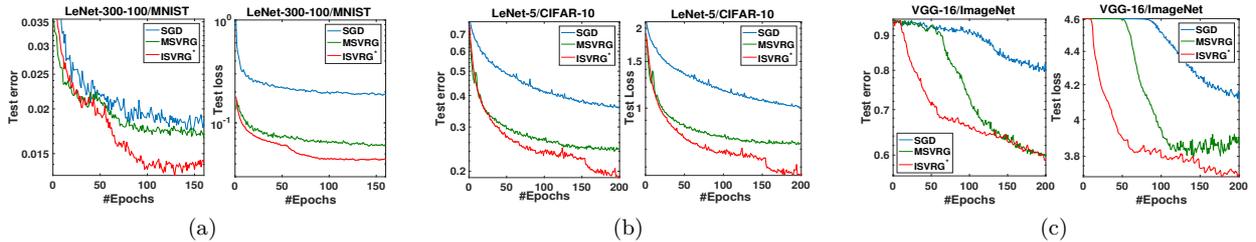

**Figure 1:** Comparison of rates of convergence in three approaches, including MSVRG, SVRG with our hybird adaptive learning rates as SVRG-$\eta_t$ and ISVRG via training/test loss. Compared with SGD and MSVRG, we can see that ISVRG* can converge faster in the beginning of epoch showed both in test error/loss on all three sub-figures, and the results significantly decrease after several epochs when the learning rate changed from decay to be fixed. Thus ISVRG* has the lowest test error/loss, which can efficiently accelerate the rates of convergence.

Consequently, when the learning rate is decayed by increasing number of iterations, a biased estimation with more reduced variance will fast converge to a point, and when the learning rate is fixed depending on the number of data samples n, a unbiased estimation with stochastic gradient will be better. All theoretical proof details are in appendix.[1]

## 5 Application

To experimentally confirm the theoretical results and insights, we train three common deep learning topologies, including LeNet[2] and VGG-16 [Simonyan and Zisserman, 2014] on three datasets including MNIST, CIFAR-10 and tiny ImageNet[3]. Our method was implemented using Caffe [4]. We use scale standard SVRG as our baseline. And we choose MSVRG method to compare with ISVRG*. Because MSVRG is a leading VR scheme based on stochastic methods which can perform better than SGD and GD for nonconvex optimization [Reddi et al., 2016b; J. Reddi et al., 2016]. To set up our experiments, a practical ways to choose the maximal value of learning rate as $\eta_t = \max\{\eta_0/ts, 1/(3Ln^{1/5})\}$, where the L = 1, 10, 100, t from 1 to m and s from 1 to the total number of epoch S. To evaluate the performance of ISVRG*, we choose the cross entropy using the softmax log loss function as the result of test error to evaluate the quality of neural networks and choose the mean squared error (MSE) that evaluate the computation cost of gradient to estimate the effectiveness of neural networks as the result of test loss.

In Figure 1, we compared the performance of SGD, MSVRG and ISVRG*. For SGD we set-up $\lambda = 0$ and $\eta = \eta_0/(ts)$ in Alg 2 and for MSVRG is $\lambda = 0.5$ and $\eta_t = \max\{\eta_0/ts, 1/(3Ln^{\frac{1}{5}})\}$ in Alg 2. In Figure 1, the blue lines showed the SGD as baseline. The red lines show the test loss and error of ISVRG* which are all lower than other methods. Particularly, the both test error and loss of ISVRG* drop down dramatically in later epochs since the adaptive learning rate start to depend on the data size n rather than t, and correspondingly the $\lambda$ change from 2/3 to 0, which means that the variance of gradients becoming high with unbiased estimator can help points escape from local minima.

## 6 Discussion

In this paper, we proposed a VR-based optimization ISVRG* for nonconvex problems. We theoretically determined that a hyper-parameter $\lambda$ working with a adaptive learning rate in each iteration can control the reduced variance of SVRG and balance the trade-off between biased/unbiased estimator. Moreover, to verify our theoretical results, we experiment on different datasets on deep learning models to estimate the range of $\lambda$ and compare these with other leading results. Both theoretical and experimental results are shown ISVRG* can efficiently accelerate rates of convergence and is faster than SVRG and SGD for nonconvex optimization.

---

[1] The file link as: https://send.firefox.com/download/207952f62e/#o28q01J-BQ7giByeKbnLLg

[2] We choose two types of LeNet, including LeNet-300-100 which has two fully connected layers as hidden layers with 300 and 100 neurons respectively, and LeNet-5 which has two convolutional layers and two fully connected layers

[3] Tiny ImageNet is a subset of ImageNet challenge (2012 ILSVRC [Russakovsky et al., 2015]), which contains 500 categories. Each category has 600 training images, 200 validation images and 200 test images, each images is re-sized to 96×96 pixels

[4] Caffe is a deep learning framework. Source code can be download: http://caffe.berkeleyvision.org